\documentclass{article}
\usepackage[T1]{fontenc} 
\usepackage[utf8]{inputenc} 
\usepackage{ismir,amsmath,cite,url}
\usepackage{graphicx}
\usepackage{color}

\usepackage{booktabs}
\usepackage{multirow}
\usepackage{pifont}
\newcommand{\cmark}{\ding{51}}%
\newcommand{\xmark}{\ding{55}}%

\usepackage{lineno}

\newcommand{\iec}{i.\,e.,\,}

\newcommand{\egc}{e.\,g.,\,}
\newcommand{\eg}{e.\,g.\,}

\newcommand{\Sect}[1]{Section~\ref{#1}}

\newcommand{\Fig}[1]{Figure~\ref{#1}}

\newcommand{\Table}[1]{Table~\ref{#1}}

\title{Learning to Read and Follow Music in \\Complete Score Sheet Images}

\multauthor
{Florian Henkel$^1$ \hspace{1cm} Rainer Kelz$^2$ \hspace{1cm} Gerhard Widmer$^{1}$} {
 $^1$ Institute of Computational Perception, Johannes Kepler University Linz, Austria\\
$^2$ Austrian Research Institute for Artificial Intelligence (OFAI), Vienna, Austria\\
{\tt\small florian.henkel@jku.at}
}
\def\authorname{F. Henkel, R. Kelz, and G. Widmer}

\begin{document}
\setlength{\textfloatsep}{11pt}
\maketitle
\begin{abstract}
This paper addresses the task of score following in sheet music
given as unprocessed images.
While existing work either relies on OMR software to obtain
a computer-readable score representation, or crucially relies on
prepared sheet image excerpts, we propose the first system that
directly performs score following in full-page, completely unprocessed
sheet images.
Based on incoming audio and a given image of the score, our system
directly predicts the most likely position within the page that
matches the audio, outperforming current state-of-the-art
image-based score followers in terms of alignment precision.
We also compare our method to an OMR-based approach
and empirically show that it can be a viable alternative to
such a system.
\end{abstract}
%

\section{Introduction}\label{sec:introduction}
Score following is a fundamental task in MIR and the basis for applications such as automatic accompaniment \cite{Raphael10_MusicPlusOneML_ICML, Cont10_ScoreMusic_IEEE-TPAMI}, automatic page turning \cite{ArztWD08_PageTurning_ECAI} or the synchronization of live performances to visualizations \cite{ArztFGGGW15_AIConcertgebouw_IJCAI, ProckupGHK13_Orchestra_IEEE-MM}. These applications require a real-time capable system that aligns a musical performance to a symbolic score representation in an online fashion.
To solve this, existing systems either require a computer-readable score representation (\eg extracted using Optical Music Recognition (OMR)\cite{CalvoHP19_UnderstandingOMR_CRR}) or rely on fixed-size (small) snippets of sheet images.

Models from the latter category are by design only capable of handling fixed-sized excerpts of the sheet image due to a limited action space to predict the next position in the score.
This is a severe constraint, as the sheet image snippet has to (at least partly) correspond to the incoming audio excerpt. If it does not match the audio anymore (due to some tracking error), no proper prediction can be formed.
To overcome this limitation, we attempt score following directly in the full sheet image, enabling the system to observe the whole page at any given time. This makes the problem significantly more challenging, \egc due to repetitive musical score structures, compared to locally constrained systems that only look at snippets.

To the best of our knowledge, we present the first system that requires neither OMR nor any other form of score pre-processing and directly follows musical performances in full sheet images in an end-to-end fashion.\footnote{Code and data will be made available on-line: \url{https://github.com/CPJKU/audio_conditioned_unet}} 

Specifically, we formulate score following as a \textit{referring image segmentation task} and introduce an appropriate model architecture in \Sect{sec:image_seg}, including a conditioning mechanism in the form of a Feature-wise Linear Modulation (FiLM) layer \cite{PerezSDVDC18_FILM_AAAI} as a central building block.
In \Sect {sec:experiments}, we demonstrate the system on polyphonic piano music, taken from the MSMD dataset \cite{DorferHAFW18_MSMD_TISMIR}. To analyze its generalization capabilities, we also test it on real musical performances taken from the MSMD test split, in \Sect{sec:real_perf}. The results will show that our model outperforms current state-of-the-art image based trackers in terms of alignment precision, but also that it currently lacks robustness across different audio conditions.

\begin{figure*}[t]
\centering
\includegraphics[width=0.78\textwidth]{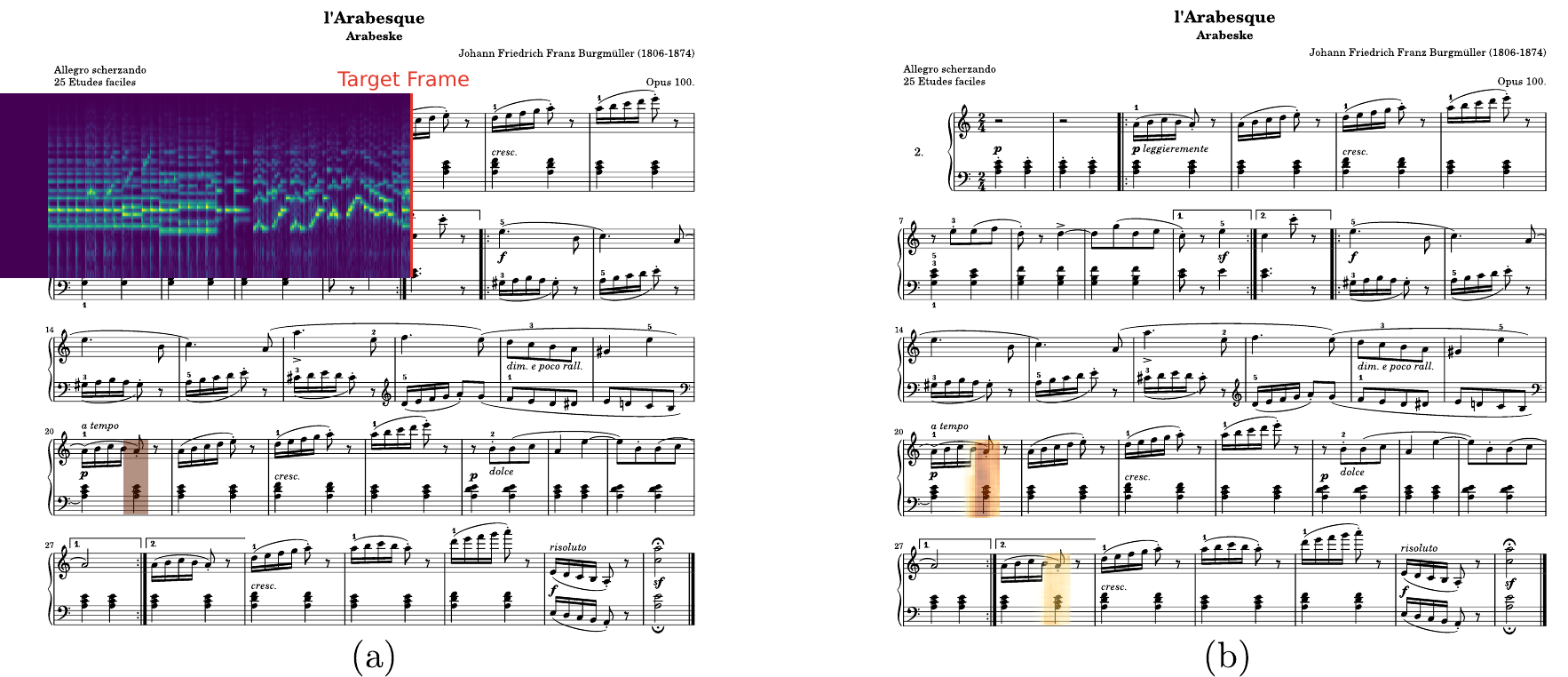}
 \caption{\small{
Score Following Task as modelled in this work: Given a score (sheet image) and an incoming musical performance (audio), the goal is to predict the current location in the score in real time. The audio is fed to the tracker frame by frame and the system processes the last 40 frames to predict the position for the latest frame (the \textit{Target Frame} marked red in (a)).
The ground truth (bounding box around current score position; see (a)) is given as a binary segmentation mask. The system predicts a probability for each pixel to correspond to the current audio; thus it highlights those regions that are most likely to match the correct location. Ideally, this should be only a single region. However, in (b) we see that such a prediction is not perfect: while the highest probability is assigned to the correct position in staff four, there is also a small likelihood in the last staff, as the notes are the same for both locations. To predict the location correctly, the system needs to consider the whole audio up to the current point in time, which motivates our design choices introduced in \Sect{sec:image_seg}.}
}
\label{fig:sf_task}
\end{figure*}


\section{Related Work}
\label{sec:related_work}
Score following approaches can be broadly categorized into those that rely on the presence of a computer-readable score representation, such as MusicXML or MIDI, and those that try to do without such a symbolic representation.
In the former category, techniques like Dynamic Time Warping (DTW)  \cite{Arzt16_MusicTracking_PhD, ArztFGGGW15_AIConcertgebouw_IJCAI, Dixon05_ODTW_IJCAI} and Hidden Markov Models (HMMs) \cite{Cont06_ScoreFollowingViaNmfAndHMM_ICASSP, OrioLS03_scorefollowing_NIME, NakamuraCCOS15_ScoreFollowingSemiHMM_ISMIR} are applied to achieve robust and reliable tracking results. The main issue with these approaches is the need for computer-readable score representations, which must either be created manually in a tedious and time consuming process, or automatically extracted using OMR. In the OMR case, the faithfulness of the symbolic score to what is depicted on the sheet image strongly depends on the quality of the OMR system, which may introduce errors that impede the actual score following task. Empirical evidence for this claim was published in \cite{HenkelBDW19_ScoreFollowingRL_TISMIR}, where a DTW-based score following system that relied on MIDI scores extracted via an OMR system had difficulties tracking synthetically created test data.

Several recent publications deal with the latter category, and investigate score following in the context of non-computer-readable score representations, represented as raw sheet images. In \cite{DorferAW16_ScoreFollowDNN_ISMIR}, the authors propose a multi-modal deep neural network to predict the position within a sheet snippet based on a short audio excerpt. In \cite{DorferHW18_ScoreFollowingAudioSheet_ISMIR} and \cite{HenkelBDW19_ScoreFollowingRL_TISMIR}, score following is formulated as a reinforcement learning (RL) problem, where the RL agent's task is to adapt its reading speed in an unrolled sheet image, conditioned on an audio excerpt.
One of the limitations of all these methods is that they require the score to be represented in an unrolled form, \iec staves need to be detected in the sheet image, cut out and presented to the score following system in sequence. 

To overcome this, \cite{HenkelKW19_AudioConditionedUNet_WORMS} introduced a system that directly infers positions within full sheet images for monophonic piano music. However, the temporal aspect
of score following was neglected altogether --- based on an audio excerpt all possible matching positions in the full sheet image are highlighted, including those that were already played --- making it interesting preliminary work, but not a reasonable score following system. 
In the following we build upon their foundation and incorporate long term audio context, proposing the first fully capable score following system that works on entire sheet images, without needing any pre-processing steps. 

\section{Score Following as a Referring Image Segmentation Task}\label{sec:image_seg}
Similarly to \cite{HenkelKW19_AudioConditionedUNet_WORMS}, we model score following as an image segmentation task --- more specifically, as a \textit{referring} image segmentation task. In computer vision, the goal of referring image segmentation is to identify a certain region or object within an image based on some language expression \cite{MaoHTCYM16_GenCompObjects_CVPR, YeRLW19_CrossModalAttention_CVPR}. It shows similar characteristics as the multi-modal approach to score following --- we want to locate the position in the sheet image that the incoming audio refers to, meaning we treat the audio as the language expression, and the score image as the entity to reason about.
More precisely, our modeling setup is as follows: based on the incoming musical performance up to the current point in time, the task of the model is to predict a segmentation mask for the given score that corresponds to this currently played music, as shown in \Fig{fig:sf_task}. The ground truth for this task is chosen to be a region around the current position in the score with a fixed width and a height depending on the height of the staff. While the size of this mask can be arbitrarily chosen, we define it such that it provides a meaningful learning target first and foremost.

A challenging question arising with such a setup is how to combine the different input modalities, audio and score. While \cite{DorferAW16_ScoreFollowDNN_ISMIR, DorferHW18_ScoreFollowingAudioSheet_ISMIR, HenkelBDW19_ScoreFollowingRL_TISMIR} learn a latent representation for both input modalities which are subsequently concatenated and further processed, we follow the direction of \cite{HenkelKW19_AudioConditionedUNet_WORMS} instead, and employ a \textit{conditioning mechanism} that directly modulates the activity of feature detectors that process the score image. 
The former setup would be problematic due to the increasing number of parameters. Furthermore, this design is able to retain the fully-convolutional property of our model, \iec if desired one could process sheet images of arbitrary resolution.\footnote{Note that while we do not investigate this further and work with fixed-sized sheets, this could be useful in a real world application.} 
In contrast to \cite{HenkelKW19_AudioConditionedUNet_WORMS}, we apply the conditioning mechanism on top of a recurrent layer to provide a longer temporal context. This permits the audio input up until the current point in time to guide the segmentation towards the corresponding position in the score image. We argue that it is necessary for this task to have such a long temporal audio context in order to form more reliable predictions.
For example, it is common to have repeating note patterns in the score spanning over a longer period of time in the audio. Existing trackers that use only a fixed-size audio input are not able to distinguish between such patterns, if they exceed the given audio context.

\subsection{Feature-wise Linear Modulation}
\begin{figure}
\centering
\includegraphics[width=1.0\columnwidth]{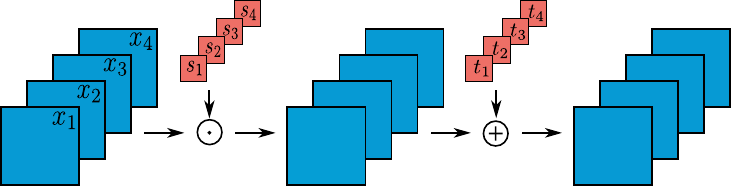}
 \caption{\small{
 Sketch of the FiLM layer \cite{PerezSDVDC18_FILM_AAAI}. The layer scales and translates the feature maps $\mathbf{x}$ using learned functions $\mathbf{s}(\mathbf{z})$ and $\mathbf{t}(\mathbf{z})$, respectively. $\mathbf{z}$ is an additional, external input carrying the conditioning information.
 }}
\label{fig:film_layer}
\end{figure}

The Feature-wise Linear Modulation (FiLM) layer is a simple linear affine transformation of feature maps, conditioned on an external input \cite{PerezSDVDC18_FILM_AAAI}. The purpose of using this layer is to directly interfere with the learned representation of the sheet image by modulating its feature maps, assisting the convolutional neural network to focus only on those parts that are required for a correct segmentation. In our case, the external input $\mathbf{z}$ is the hidden state of a recurrent layer that takes as input an encoded representation of an audio excerpt. This encoded representation is created by a neural network, \egc as depicted in \Table{tab:spec_net}.
The FiLM layer itself is defined as
\begin{equation}
    f_{\text{FiLM}}(\mathbf{x}) = \mathbf{s}(\mathbf{z}) \cdot \mathbf{x} + \mathbf{t}(\mathbf{z}),
\end{equation}
where $\mathbf{s}(\cdot)$ (for scaling) and $\mathbf{t}(\cdot)$ (for translation) are arbitrary vector-valued functions implemented as neural networks. Their values depend on the conditioning vector $\mathbf{z}$, and together they define an affine transform of the tensor $\mathbf{x}$ which refers to the collection of feature maps of a particular convolutional layer, after normalization. The affine transformation is performed per feature map, meaning that each feature map $k$ is scaled and translated by $s_k(\cdot)$ and $t_k(\cdot)$, respectively, with $k$ identifying the $k$-th output of the two functions. The number of output values for $\mathbf{s}(\cdot)$ and $\mathbf{t}(\cdot)$ is the same as the number of feature maps contained in $\mathbf{x}$, denoted by $K$ (cf. \Fig{fig:film_layer}).

\subsection{Model Architecture}\label{sec:model_architecture}

\begin{figure*}[t]
\centering
\includegraphics[width=0.80\textwidth]{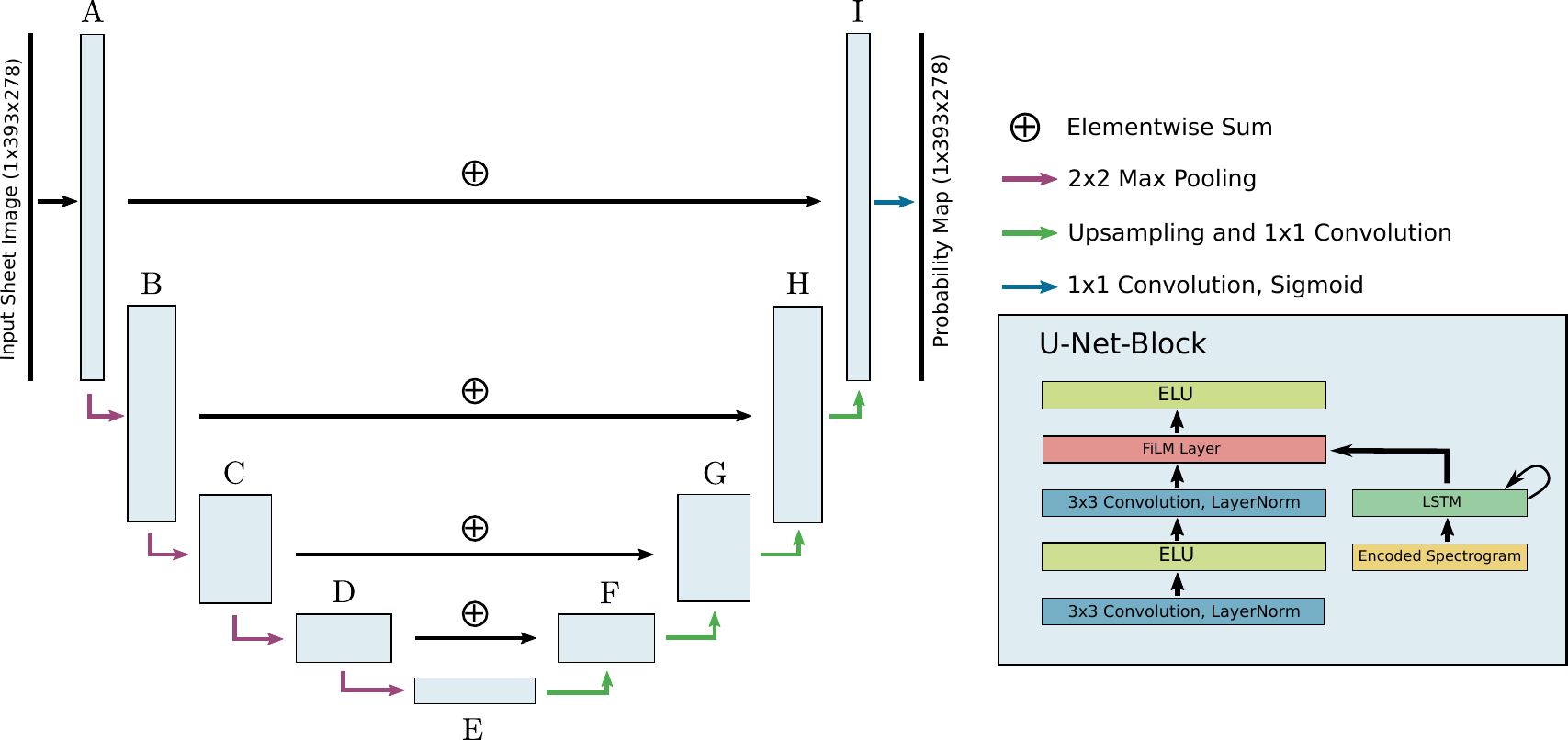}
 \caption{\small{Audio-Conditioned U-Net architecture. Each block (A-I) consists of two convolutional layers with ELU activation and layer normalization.
 The FiLM layer is placed before the last activation function. The spectrogram encoding given by the output of the network shown in \Table{tab:spec_net} is passed through a recurrent layer. The hidden state of this recurrent layer is then used for conditioning in the FiLM layer. Each symmetric block has the same number of filters starting with 8 in block A and increasing with depth to 128 in block E.}}
\label{fig:architecture}
\end{figure*}

\begin{table}
 \centering
  \small
 \begin{tabular}{cc}
 \toprule
\textbf{Audio (Spectrogram)}  $78 \times40$ \\
 \midrule
2 x ( Conv(3, 1, 1)-24 - LN - ELU ) - MP(2) \\
2 x ( Conv(3, 1, 1)-48 - LN - ELU ) - MP(2) \\
2 x ( Conv(3, 1, 1)-96 - LN - ELU ) - MP(2) \\
2 x ( Conv(3, 1, 1)-96 - LN - ELU ) - MP(2) \\
Conv(1, 0, 1)-96 - LN - ELU \\
Dense(32) - LN - ELU\\
\midrule
\end{tabular}
 \caption{\small{The context-based encoder used for the experiments. Conv($f$, $p$, $s$)-$k$ denotes a convolutional layer with $k$ $f \times f$ kernels, padding of $p$ and stride $s$.  We use layer normalization  (LN) \cite{BAKH16_LayerNorm_arxiv}, ELU activation \cite{ClevertUH15_ELU_ICLR} and max pooling (MP) with a pool size of $2 \times 2$. The output of the last layer is fed into a LSTM as shown in \Fig{fig:architecture}. The network resembles the one used in \cite{DorferHAFW18_MSMD_TISMIR}.}}
 \label{tab:spec_net}
\end{table}

Our model is based on a U-Net architecture similar to the one used in \cite{HajicDWP18_OMRUNET_ISMIR}~for detecting musical symbols in sheet images.
U-Nets were originally introduced for medical image segmentation, to segment an image into different parts, by classifying each pixel into either foreground or background \cite{RonnebergerFB15_UNET_MICCAI}. This fits naturally to
our interpretation of the score following task, as a process of segmenting the sheet image into a region that corresponds to the current position in the audio input and labelling everything else as background.
The overall architecture, shown in \Fig{fig:architecture}, resembles the one proposed in \cite{HenkelKW19_AudioConditionedUNet_WORMS}, with several important differences. Based on the empirical findings reported in \cite{HenkelKW19_AudioConditionedUNet_WORMS}, we decide to incorporate conditioning information from the audio input in blocks B-H, leaving only blocks A and I without conditioning. However, we substitute the transposed convolutions in the decoder part of the network with bilinear upsampling operations with a factor of two, followed by a $1\times1$ convolution, both aimed at reducing checkerboard artifacts in the segmentation \cite{OdenaDO16_DeConvCheckerboard_DISTILL}. Due to the small batch size used during training (cf. \Sect{sec:experimental_setup}) as well as the presence of the recurrent layer, we replace batch normalization with layer normalization \cite{BAKH16_LayerNorm_arxiv}.

For deriving the conditioning information from the audio input, we test two different spectrogram encoders. One takes a spectrogram snippet with a length of 40 frames, corresponding to two seconds of audio; the spectrogram is processed by the network shown in \Table{tab:spec_net}, which is roughly similar to the one used in \cite{DorferHAFW18_MSMD_TISMIR}. The other version takes as input a \textit{single spectrogram frame}, using a dense layer with 32 units, layer normalization and ELU activation function. The output of the encoders is fed to an
LSTM \cite{HochreiterS97_LSTM_NeuralComp} layer with 128 units 
and its hidden state is then used as the external input $\mathbf z$ in the FiLM layers.


\section{Experiments}\label{sec:experiments}

To study the properties of our proposed approach, we conduct experiments on polyphonic piano music provided by the MSMD \cite{DorferHAFW18_MSMD_TISMIR} dataset. While this section is mainly concerned with comparing data augmentation and different architectures, later on in \Sect{sec:real_perf} we will investigate the generalization capabilities of the system in terms of 16 real piano recordings from the MSMD test split. We will also contextualize the proposed system with related work described in \cite{HenkelBDW19_ScoreFollowingRL_TISMIR}, which we use as baselines for comparison.

\subsection{Data}
We use the \emph{Multi-modal Sheet Music Dataset (MSMD)} \cite{DorferHAFW18_MSMD_TISMIR}, a standard dataset for such evaluations, comprising polyphonic piano music from various composers including Bach, Mozart, and Beethoven. The sheet music is typeset with Lilypond\footnote{\url{http://lilypond.org/}}
and the audio tracks are synthesized from MIDI using Fluidsynth\footnote{\url{http://www.fluidsynth.org/}} together with a piano sound font.
The original MSMD splits used by \cite{HenkelBDW19_ScoreFollowingRL_TISMIR} encompass 354 train, 19 validation and 94 test pieces. The precise alignments between audio and sheet music in this dataset are created automatically. Despite that, it turned out that some of the pieces still contain alignment errors. We manually identified and fixed most of these errors, including wrongly aligned notes and missing or wrongly detected staves. One piece from the train set was removed, because we were not able to fix it. Thus, the cleaned dataset consists of 353 train, 19 validation, and 94 test pieces, which will be made publicly available.
 If a piece consists of several pages, each page is treated as a single piece and the original MIDI information is partitioned accordingly.\footnote{This is mainly done to facilitate the training procedure. In an application, this could be solved by 
 some simple `hack' that turns pages when the score follower reaches the end of a page.}
Altogether, we have 945 train, 28 validation and 125 test pages.
The rendered score images have a resolution of $1181\times835$ pixels, are downscaled by a factor of three to $393\times278$ pixels, and are used as the input to the U-Net. Preliminary tests showed that the downscaling does not significantly impact the performance, and benefits the speed of the training process. For the ground truth annotations, we rely on the automatic notehead alignment described in \cite{DorferHAFW18_MSMD_TISMIR}.
The notehead alignments yield $(x,y)$ coordinate pairs in the sheet image, which are further adjusted for our purposes such that the $y$ coordinates correspond to the middle of the staff the respective note belongs to. Given these coordinates, we create a binary mask  with a width of 10 pixels and an adaptive height depending on the height of the staff (see \Fig{fig:sf_task}). The task of the U-Net is now to infer a segmentation mask given the image of the score together with the conditioning information derived from the audio input. Note that in theory it should be possible to directly predict $x$ and $y$ coordinates instead of a segmentation mask, however as shown in \cite{LiuLMPFSY18_CoordConv_NeurIPS} this is a much harder task, and we were not able to achieve acceptable performance so far, even using their proposed \emph{CoordConv} layer.

The audio is sampled with 22.05~kHz and processed at a frame rate of 20 frames per second. The DFT is computed for each frame with a window size of 2048 samples and then transformed with a semi-logarithmic filterbank that processes frequencies between 60~Hz and 6~kHz, yielding 78 log-frequency bins. Lastly, the spectrogram bins are standardized to zero mean and unit variance. The audio conditioning network is presented either with 40 consecutive frames (two seconds of audio) or a single frame at a time. We use the \verb|madmom| python library
for all signal processing related computations \cite{BoeckKSKW16_Madmom_ACMMM}.

\subsection{Baselines and Evaluation Measures}
\label{sec:baselines}

In the following, we will present a series of experiments, comparing the new proposed full-page tracking system to several baselines (in order to better understand the importance of some of our design choices) as well as to related state-of-the-art approaches from the literature.

First, we evaluate two different spectrogram encoders, as introduced in \Sect{sec:model_architecture}, vis-\`a-vis a baseline version of our system that does not have the capability to summarize all the audio up to the current point in time, \iec that does not have memory in the form of an RNN. We do this in order to obtain empirical evidence for our argument that having access to long term temporal information is highly beneficial for obtaining good approximate solutions to the score following task. The two different encoders are denoted as context-based (CB) and frame-based (FB), using 40 spectrogram frames and a single frame, respectively. The baseline without temporal context uses the CB encoder and replaces the RNN layer with a fully connected layer of the same size. In the following this baseline will be denoted as NTC (no temporal context). 

The \textit{evaluation measures} used for this comparison are of a geometric kind (bounding box pixel error and distance on printed score page), in order to focus on the new challenge of full-page orientation: we measure the pixel-wise evaluation metrics \emph{Precision}, \emph{Recall} and $F_1$-score that were also used in \cite{HenkelKW19_AudioConditionedUNet_WORMS}, and the mean and median alignment error between ground truth and prediction in centimeters, both with the network output thresholded at 0.5. To calculate the alignment error between the ground truth and the predicted probability mask (recall \Fig{fig:sf_task}), we calculate the center of mass over all pixels for both masks and compute the euclidean distance between the two centers to obtain the alignment error in pixels. Given a resolution of 72 dpi, the error is converted to centimeter using a factor of 0.0352 cm/pixel, under the assumption that the score image is printed on a sheet of DIN A4 paper.

In the second experiment we compare our system to alternative state-of-the-art approaches from the literature:
the first approach is based on an OMR system that extracts symbolic MIDI information from the sheet image. The symbolic MIDI information is then synthesized to audio. The system subsequently computes chroma features with a feature rate of 20~Hz from both the synthesized and the performance audio, and applies audio-to-audio alignment using a form of online DTW \cite{Mueller15_FMP_SPRINGER}. 
This is the method described in \cite{HenkelBDW19_ScoreFollowingRL_TISMIR} and will be abbreviated as OMR in the upcoming result table.
The second and third approach, described in \Sect{sec:related_work}, are a multi-modal localization network (MM-Loc) \cite{DorferAW16_ScoreFollowDNN_ISMIR} and a Reinforcement Learning (RL) agent \cite{DorferHW18_ScoreFollowingAudioSheet_ISMIR, HenkelBDW19_ScoreFollowingRL_TISMIR}, both working with sheet image snippets.

The \textit{evaluation measure} for this will be of a more music-related kind (temporal tracking error in the performance), reflecting the intended purpose of the systems (score following), and permitting a direct comparison with alternative methods. Similarly to \cite{Dixon05_ODTW_IJCAI, Arzt16_MusicTracking_PhD}, we compute, for each note onset, the absolute time difference between prediction and ground truth. We set 5 threshold values, ranging from 0.05 to 5 seconds, and report the cumulative percentage of notes tracked with an error up to the given threshold.
Given the ground truth alignment from note onsets to the corresponding notehead coordinates in the sheet image, we can interpolate from the predicted positions in the sheet image back to the time domain. 
This is straightforward for MM-Loc and the RL agent, because they both already use an unrolled score derived from the groundtruth, whereas the proposed method requires further processing.
We first need to compute the center of mass of the segmented region to obtain $x,y$ coordinates. We map the $y$ coordinate to the closest staff, and apply a similar interpolation as before in an unrolled score to get the time difference between the predicted and actual position in the score.

For evaluating the OMR baseline we face a problem that has already been noted in \cite{HenkelBDW19_ScoreFollowingRL_TISMIR} --- we do not have the required groundtruth alignment between the OMR-extracted score and the performance. Given that only onset positions are evaluated, we are justified to assume a perfect alignment between score and audio, if for each unit of time in the audio a constant distance in the score sheet is travelled. If the OMR system makes no errors, the alignment between OMR score and performance is a diagonal in the DTW global cost matrix, correcting the overall tempo difference by a linear factor. As in \cite{HenkelBDW19_ScoreFollowingRL_TISMIR}, we evaluate the OMR-based system by measuring the offset of the actual tracking position relative to the perfect alignment.

\subsection{Experimental Setup} \label{sec:experimental_setup}

All models are trained using the same procedure.
We optimize the \emph{Dice} coefficient loss \cite{MilletariNA16_VNet_3DV}, which is more suitable than \egc \emph{binary cross-entropy}, as we are facing an imbalanced segmentation problem with far more unimportant background pixels than regions of interest.
To optimize this target we use Adam \cite{KingmaB14_Adam_ICLR} with default parameters, an initial learning rate of $1e^{-4}$ and $L^2$ weight decay with a factor of $1e^{-5}$. If the conditioning architecture involves an LSTM, we use a batch-size of 4 and a sequence length of 16. For the audio conditioning model without a temporal context we use a batch size of 64. The weights of the network are initialized orthogonally \cite{SaxeMG13_OrthogonalInit_arxiv} and the biases are set to zero. If the loss on the validation set does not decrease for 5 epochs, we halve the learn rate and stop training altogether when the validation loss does not decrease over a period of 10 epochs or the maximum number of 100 epochs is reached. The model parameters with the lowest validation loss are used for the final evaluation on the test set.  Similar to \cite{HenkelKW19_AudioConditionedUNet_WORMS}, we perform data augmentation in the image domain by shifting the score images along the $x$ and $y$ axis.
To investigate whether tempo augmentation improves model performance, we train all models without tempo augmentation as well as with 7 different tempo change factors ranging from 0.5 up to 1.5.

\subsection{Results}

In \Table{tab:architecture_comparison}, we compare different conditioning architectures, no long term temporal context (NTC), a context of 40 frames (CB) and a single frame (FB) in combination with an LSTM, respectively. We observe that the NTC model has the lowest performance, both in terms of the pixel-wise measures, as well as in terms of its alignment error. A possible reason for this could be ambiguities in the sheet image, since audio excerpts could match several positions in the score.
The results for CB and FB support our initial claim that a long term temporal context is required for this task. While both models achieve a good performance, CB outperforms FB in all measures. On average, the alignment error is around 1.25 cm and the median is at 0.51 cm, meaning that half of the time our model is less than 0.51 cm away from the true position. Furthermore, we observe that tempo augmentation improves the results for all models.

In \Table{tab:method_comparison}, we compare our best model from \Table{tab:architecture_comparison} to several baselines from the literature in terms of the cumulative percentage of onsets that are tracked with an error below a given threshold. We observe that the context-based proposed model (CB) outperforms all baselines except for the highest threshold. This suggests that our method is very precise on one hand, but on the other hand is not able to track all onsets with a timing error below five seconds. 

\begin{table}
\centering
\small
\begin{tabular}{lcccccc}

\multicolumn{7}{c}{MSMD (125 test pages)}             \\\toprule
      & \textbf{TA} & \textbf{P} & \textbf{R} & $\mathbf{F_1}$ & $\mathbf{\overline{d}_{cm}}$ & $\mathbf{\widetilde{d}_{cm}}$ \\\midrule
\multirow{2}{*}{NTC} & \xmark & 0.696 & 0.665 & 0.678 & 3.70 & 2.37\\
                     & \cmark & 0.770 & 0.740 & 0.754 & 2.78 & 1.61\\\midrule
                                   
\multirow{2}{*}{CB}& \xmark & 0.810 & 0.790 & 0.799 & 1.62 & 0.73 \\
                   & \cmark & \textbf{0.854} & \textbf{0.835} & \textbf{0.843} & \textbf{1.25} & \textbf{0.51} \\\midrule
\multirow{2}{*}{FB}  & \xmark & 0.790 & 0.768 & 0.778 & 1.82 & 1.21 \\
                     & \cmark & 0.820 & 0.816 & 0.816 & 1.58 & 0.80 \\\midrule
\end{tabular}
\caption{\small{ Different conditioning architectures with/without tempo augmentation (TA): no temporal context (NTC), context-based (CB) and frame-based (FB). For each model the parameters with lowest validation loss are chosen for evaluation on the test set. Measures: pixel-wise precision (P), recall (R) and $F_1$, and mean ($\overline{d}_{cm}$) and median ($\widetilde{d}_{cm}$) of alignment error in centimeters.
}}
 \label{tab:architecture_comparison}
\end{table}

\begin{table}
\centering
\small
\begin{tabular}{lcccc}
  \multicolumn{5}{c}{MSMD (125 test pages)}             \\\toprule
  \textbf{Err. [sec]} & OMR\cite{HenkelBDW19_ScoreFollowingRL_TISMIR} & MM-Loc \cite{Dorfer18_MusicTracking_PhD} & RL\cite{HenkelBDW19_ScoreFollowingRL_TISMIR} & CB \\\midrule
  
  $\leq 0.05$ & 44.7\% & 44.6\% & 40.9\% & \textbf{73.3\%} \\
  $\leq 0.10$ & 51.9\% & 49.2\% & 43.3\% & \textbf{74.7\%} \\
  $\leq 0.50$ & 76.0\% & 82.2\% & 79.7\% & \textbf{85.2\%} \\
  $\leq 1.00$ & 85.0\% & 86.0\% & 87.8\% & \textbf{88.5\%} \\
  $\leq 5.00$ & \textbf{97.4\%} & 92.0\% & 97.2\% & 93.7\%\\\midrule
\end{tabular}
\caption{\small{ Our best model (CB)
vs.~existing baselines, in terms of onsets tracked with an error below a given threshold. 
For the RL agent we report the average over 10 runs due to its stochastic policy. In contrast to \cite{HenkelBDW19_ScoreFollowingRL_TISMIR}, OMR, MM-Loc and RL do not stop tracking if they fall out of a given tracking window. }}
\label{tab:method_comparison}
\end{table}

\section{Real Performances}\label{sec:real_perf}

To test the generalization capabilities of the system under real recording conditions, we evaluate our best model on the 16 piano recordings (corresponding to 25 score pages) from the MSMD test split introduced in \cite{HenkelBDW19_ScoreFollowingRL_TISMIR}, for which we also manually corrected some of the alignments.
We compare again to the baselines introduced in \Sect{sec:baselines}, which are likewise evaluated using the corrected alignments.
In line with \cite{HenkelBDW19_ScoreFollowingRL_TISMIR}, we compare four different settings with increasing difficulty.
The first is the same synthetic setting as in \Sect{sec:experiments}.
The second setting uses the performance MIDI synthesized with the same piano synthesizer used during training. The third uses the audio of the ``direct out'' audio output of the ``Yamaha AvantGrand N2'' hybrid piano used for recording, and the last one uses the audio recorded via a room microphone. 

\Table{tab:eval_methods_rp} summarizes the results. Overall, we observe that the proposed system (CB) achieves more precise results in terms of time difference (i.e., higher percentages for the tighter error thresholds) in three out of four settings. For the last setting we observe a worse performance, which indicates that our model has possibly overfit to the synthesized audio and is not yet robust enough. OMR yields very robust results in all scenarios, which is possibly due to the used chroma features. While the results are not as precise, it outperforms the other methods for higher threshold values. 

A possible explanation for this is that our model has more freedom in being able to perform big jumps on the sheet image paper, thus increasing the error possibility. Models relying on sheet snippets are not designed to perform such jumps and thus can also not make very extreme errors.
Furthermore, our model is more sensitive to the audio representation fed into the conditioning mechanism, as it influences the convolutional filters in multiple layers that process the sheet image. 
Overall, we assume that this is an issue of the synthetic dataset which can be tackled by training on more diverse performances and a more robust audio model for the conditioning mechanism.

\begin{table}[ht]
\centering
\small
\begin{tabular}{lcccc}
 \toprule
\textbf{Err. [sec]} & OMR \cite{HenkelBDW19_ScoreFollowingRL_TISMIR} & MM-Loc  \cite{Dorfer18_MusicTracking_PhD} & RL  \cite{HenkelBDW19_ScoreFollowingRL_TISMIR} & CB \\\midrule
 \multicolumn{5}{r}{Original MIDI Synthesized (Score = Performance)} \\
 \midrule
  $\leq 0.05$ & 37.1\% & 41.6\% & 36.5\% & \textbf{69.8\%}  \\
  $\leq 0.10$ & 46.1\% & 44.2\% & 38.2\% & \textbf{70.6\%} \\
  $\leq 0.50$ & 74.9\% & 77.6\% & 72.9\% & \textbf{80.6\%}  \\
  $\leq 1.00$ & \textbf{86.8\%} & 79.9\% & 79.8\% & 82.4\%\\
  $\leq 5.00$ & \textbf{99.6\%} & 90.3\% & 96.5\% & 89.1\%  \\
 \midrule
 \multicolumn{5}{r}{Performance MIDI Synthesized} \\
 \midrule
  $\leq 0.05$ & 28.9\% & 47.2\% & 23.4\%  & \textbf{56.5\%}  \\
  $\leq 0.10$ & 39.8\% & 49.0\% & 24.8\%  & \textbf{58.1\%} \\
  $\leq 0.50$ & 71.7\% & \textbf{83.2\%} & 54.5\%  & 80.9\% \\
  $\leq 1.00$ & 83.4\% & \textbf{86.1\%} & 64.0\%  & 84.4\% \\
  $\leq 5.00$ & \textbf{98.8\%} & 96.0\% & 81.2\%  & 90.1\%  \\
 \midrule
 \multicolumn{5}{r}{Direct Out} \\
 \midrule
  $\leq 0.05$ & 22.6\% & 33.8\% & 27.7\%  & \textbf{40.0\%}  \\
  $\leq 0.10$ & 33.0\% & 35.4\% & 29.1\%  & \textbf{41.6\%}  \\ 
  $\leq 0.50$ & \textbf{70.3\%} & 59.7\% & 60.7\%  & 64.2\% \\
  $\leq 1.00$ & \textbf{83.9\%} & 63.4\% & 73.3\%  & 69.3\% \\
  $\leq 5.00$ & \textbf{99.3\%} & 75.3\% & 95.5\%  & 81.1\%  \\
  \midrule
 \multicolumn{5}{r}{Room Recording} \\
 \midrule
  $\leq 0.05$ & \textbf{22.6\%}  & 20.7\% & 19.2\% & 9.4\%   \\
  $\leq 0.10$ & \textbf{32.2\%}  & 24.3\%  & 20.6\% & 10.5\%   \\
  $\leq 0.50$ & \textbf{70.2\%}  & 54.1\%  & 46.6\% & 21.5\%    \\
  $\leq 1.00$ & \textbf{82.7\%}  & 57.3\%  & 58.7\% & 26.2\%    \\
  $\leq 5.00$ & \textbf{97.4\%}  & 70.2\%  & 89.1\% & 44.3\%   \\
 \bottomrule
 \end{tabular}
\caption{\small{ Comparing best performing model to several baselines on a set of 16 real piano recordings (25 pages) from the MSMD test split. Model evaluation is as described in \Table{tab:method_comparison}, with the difference that for the RL agent we report the average over 50 runs due to its stochastic policy and the smaller sample size.}}
 \label{tab:eval_methods_rp}
\end{table}

\section{Discussion and Conclusion}\label{sec:conclusion}
We have proposed the first end-to-end trained score following system that directly works on full sheet images.
The system is real-time capable due to a constant runtime per step, it compares favorably with existing baselines on synthetic polyphonic piano music, and sets the new state of the art for sheet-image-based score following in terms of temporal alignment error.
However, there are still generalization problems for real piano recordings.
While the model shows a much more precise alignment in most scenarios, we see a performance deterioration over different recording conditions. 
This will need to be solved in the future, either with a more robust audio model, or a data augmentation strategy that incorporates reverberation effects.
Future work will also require testing on scanned or photographed sheet images, to gauge generalization capabilities of the 
system in the visual domain as well. As there is currently no dataset consisting of scanned sheet images with precise notehead to audio alignments, it will be necessary to curate a test set. 
The next step towards a system with greater capabilities, is to either explicitly or implicitly incorporate a mechanism to handle repetitions in the score as well as in the performance.
We assume that the proposed method will be able to acquire this capability quite naturally from properly prepared training data, although we suspect its performance will heavily depend on its implicit encoding of the audio history so far, \iec how large an auditory context the recurrent network is able to store.

\section{Acknowledgments}
This project has received funding from the European Research Council (ERC) 
under the European Union's Horizon 2020 research and innovation program
(grant agreement number 670035, project "Con Espressione"). 

\bibliography{main}

\end{document}